\documentclass{article}

\usepackage[margin=1.2in]{geometry}
\usepackage[utf8]{inputenc} 
\usepackage[T1]{fontenc}    
\usepackage{hyperref}       
\usepackage{url}            
\usepackage{booktabs}       
\usepackage{amsfonts}       
\usepackage{nicefrac}       
\usepackage{microtype}      

\usepackage{algpseudocode}
\usepackage{algorithm}
\usepackage{dblfloatfix}
\usepackage{subcaption}
\usepackage{float}
\usepackage{makecell}
\usepackage{graphicx}

\title{Unrolled Optimization with Deep Priors}

\author{Steven Diamond\thanks{These authors contributed equally.} \and Vincent Sitzmann\footnotemark[1] \and Felix Heide \and Gordon Wetzstein
}

\begin{document}

\maketitle

\begin{abstract}
	A broad class of problems at the core of computational imaging, sensing, and
  low-level computer vision reduces to the inverse problem of extracting latent
  images that follow a prior distribution, from measurements taken under a known
  physical image formation model.
  Traditionally, hand-crafted priors along with iterative optimization methods
  have been used to solve such problems.
	In this paper we present unrolled optimization with deep priors,
  a principled framework for infusing knowledge of the image formation
  into deep networks that solve inverse problems in imaging,
  inspired by classical iterative methods.
  We show that instances of the framework outperform the state-of-the-art by a substantial margin for a wide variety of imaging problems, such as denoising, deblurring, and compressed sensing magnetic resonance imaging (MRI). 
  Moreover, we conduct experiments that explain how the framework 
  is best used and why it outperforms previous methods. 
\end{abstract}

\newcommand{\vect}[1]{\mathbf{#1}}
\newcommand{\mat}[1]{\mathbf{#1}}

\newcommand{\reals}{{\mbox{\bf R}}}
\newcommand{\integers}{{\mbox{\bf Z}}}
\newcommand{\complex}{{\mbox{\bf C}}}

\newcommand\comment[1]{\textcolor{red}{#1}}

\newcommand{\psf}{k}

\newcommand{\argmin}[1]{\stackrel[\{ #1 \}]{}{\textrm{arg min}}}
\newcommand{\minimize}[1]{\stackrel[\{ #1 \}]{}{\textrm{minimize}}}

\newcommand{\BEAS}{\begin{eqnarray*}}
\newcommand{\EEAS}{\end{eqnarray*}}
\newcommand{\BEA}{\begin{eqnarray}}
\newcommand{\EEA}{\end{eqnarray}}
\newcommand{\BEQ}{\begin{equation}}
\newcommand{\EEQ}{\end{equation}}
\newcommand{\BIT}{\begin{itemize}}
\newcommand{\EIT}{\end{itemize}}
\newcommand{\BNUM}{\begin{enumerate}}
\newcommand{\ENUM}{\end{enumerate}}

\newcommand{\grad}{\nabla}

\newcommand{\BA}{\begin{array}}
\newcommand{\EA}{\end{array}}

\newcommand{\eg}{{\it e.g.}}
\newcommand{\ie}{{\it i.e.}}
\newcommand{\etc}{{\it etc.}}
\newcommand{\etal}{{\it et al.}}

\newcommand{\ones}{\mathbf 1}
\newcommand{\Vect}{\mathop{\bf vec}}
\newcommand{\Argmin}{\mathop{\rm argmin}}
\newcommand{\Argmax}{\mathop{\rm argmax}}
\newcommand{\prox}[1]{\mathbf{prox}_{#1}}

\section{Introduction}
In inverse imaging problems, we seek to reconstruct a latent image from
measurements taken under a known physical image formation.
Such inverse problems arise throughout computational photography, computer vision, medical imaging, and
scientific imaging.
Residing in the early vision layers of every autonomous vision system,
they are essential for all vision based autonomous agents.
Recent years have seen tremendous progress in both classical and deep methods for
solving inverse problems in imaging.
Classical and deep approaches have relative advantages and disadvantages.
Classical algorithms, based on formal optimization, exploit knowledge of the image
formation model in a principled way,
but struggle to incorporate sophisticated learned models of natural images.
Deep methods easily learn complex statistics of natural images,
but lack a systematic approach to incorporating prior knowledge of the
image formation model. 
What is missing is a general framework for designing deep networks that
incorporate prior information,
as well as a clear understanding of when prior information is useful.

In this paper we propose unrolled optimization with deep priors (ODP):
a principled, general purpose framework for integrating prior knowledge into deep networks.
We focus on applications of the framework to inverse problems in imaging.
Given an image formation model and a few generic, high-level design choices,
the ODP framework provides an easy to train, high performance 
network architecture.
The framework suggests novel network architectures that outperform prior work
across a variety of imaging problems.

The ODP framework is based on unrolled optimization,
in which we truncate a classical iterative optimization
algorithm and interpret it as a deep network.
Unrolling optimization has been a common practice among practitioners in
imaging,
and training unrolled optimization models has recently been explored for various
imaging applications, all using variants of field-of-experts priors
\cite{gregor2010learning,schmidt2014shrinkage,chen2015learning,nips2016_6074}.
We differ from existing approaches in that we propose a general framework for unrolling optimization methods along with deep convolutional prior architectures \emph{within} the unrolled optimization.
By training deep CNN priors within unrolled optimization architectures,
instances of ODP outperform state-of-the-art results on a broad
variety of inverse imaging problems.

Our empirical results clarify the benefits and limitations of encoding
prior information for inverse problems in deep networks.
Layers that (approximately) invert the image formation operator are useful
because they simplify the reconstruction task to denoising and correcting artifacts
introduced by the inversion layers. On the other hand, prior layers improve network generalization,
boosting performance on unseen image formation operators.
For deblurring and compressed sensing MRI, we found that a single ODP model trained on many image formation operators outperforms existing state-of-the-art methods where a specialized model
was trained for each operator.

Moreover, we offer insight into the open question of what iterative algorithm is best
for unrolled optimization, given a linear image formation model.
Our main finding is that simple primal algorithms that (approximately) invert
the image formation operator each iteration perform best.

In summary, our contributions are as follows:
\begin{enumerate}
  \item We introduce ODP, a principled, general purpose framework for inverse
    problems in imaging, which incorporates prior knowledge of the image formation into deep networks.
  \item We demonstrate that instances of the ODP framework for
    denoising, deblurring, and compressed sensing MRI outperform
    state-of-the-art results by a large margin.
  \item We present empirically derived insights on how the ODP 
    framework and related approaches are best used,
    such as when exploiting prior information is advantageous and
    which optimization algorithms are most suitable for unrolling.
\end{enumerate}

\section{Motivation}\label{s-background}
\paragraph{Bayesian model}
The proposed ODP framework is inspired by an extensive body of work on solving
inverse problems in imaging via maximum-a-posteriori (MAP) estimation under a Bayesian model.
In the Bayesian model, an unknown image $x$ is drawn from a prior distribution
$\Omega(\theta)$ with parameters $\theta$.
The imaging system applies a linear operator $A$ to this image, representing all optical processes in the capture,
and then measures an image $y$ on the sensor, drawn from a noise distribution $\omega(Ax)$ that models sensor noise, \eg, read noise, and noise in the signal itself, \eg, photon-shot noise.

Let $P(y | Ax)$ be the probability of sampling $y$ from $\omega(Ax)$ and
$P(x;\theta)$ be the probability of sampling $x$ from $\Omega(\theta)$.
Then the probability of an unknown image $x$ yielding an observation $y$
is proportional to $P(y|Ax)P(x;\theta)$.

The MAP point-estimate of $x$ is given by
$x = \Argmax_x P(y|Ax)P(x;\theta)$,
or equivalently
\BEQ\label{p-bayes}
  x = \Argmin_x f(y, Ax) + r(x, \theta),
\EEQ
where the data term $f(y, Ax) = -\log P(y | Ax)$ and prior term $r(x,\theta) = -\log P(x;\theta)$
are negative log-likelihoods. Computing $x$ thus involves solving an optimization problem \cite[Chap.~7]{boyd2004cvx}.

\paragraph{Unrolled iterative methods}
A large variety of algorithms have been developed for solving problem~(\ref{p-bayes})
efficiently for different convex data terms and priors
(\eg, FISTA~\cite{doi:10.1137/080716542}, Chambolle-Pock~\cite{chambolle2011first}, ADMM~\cite{Boyd:2011}).
The majority of these algorithms are iterative methods,
in which a mapping $\Gamma(x^k, A, y, \theta) \to
x^{k+1}$ is applied repeatedly to generate a series of iterates that converge
to solution $x^\star$, starting with an initial point $x^0$.

Iterative methods are usually terminated based on a stopping condition that
ensures theoretical convergence properties.
An alternative approach is to execute a pre-determined number of iterations
$N$, in other words unrolling the optimization algorithm. This approach is motivated by the fact that for many imaging applications very high accuracy, \eg, convergence below tolerance of $10^{-6}$ for every local pixel state, is not needed in practice, as opposed to optimization problems in, for instance, control. 
Fixing the number of iterations allows us to view the iterative method as an
explicit function $\Gamma^N(\cdot, A, y, \theta) \to x^N$ of the initial point $x^0$.
Parameters such as $\theta$ may be fixed across all iterations or vary by
iteration.
The unrolled iterative algorithm can be interpreted as a deep network \cite{nips2016_6074}. 


\paragraph{Parameterization}
The parameters $\theta$ in an unrolled iterative algorithm are the algorithm
hyperparameters, such as step sizes, and model parameters defining the prior.
Generally the number of algorithm hyperparameters is small (1-5 per iteration),
so the model capacity of the unrolled algorithm is primarily determined by the
representation of the prior.

\begin{figure*}
\centering
\includegraphics[width=0.7\textwidth]{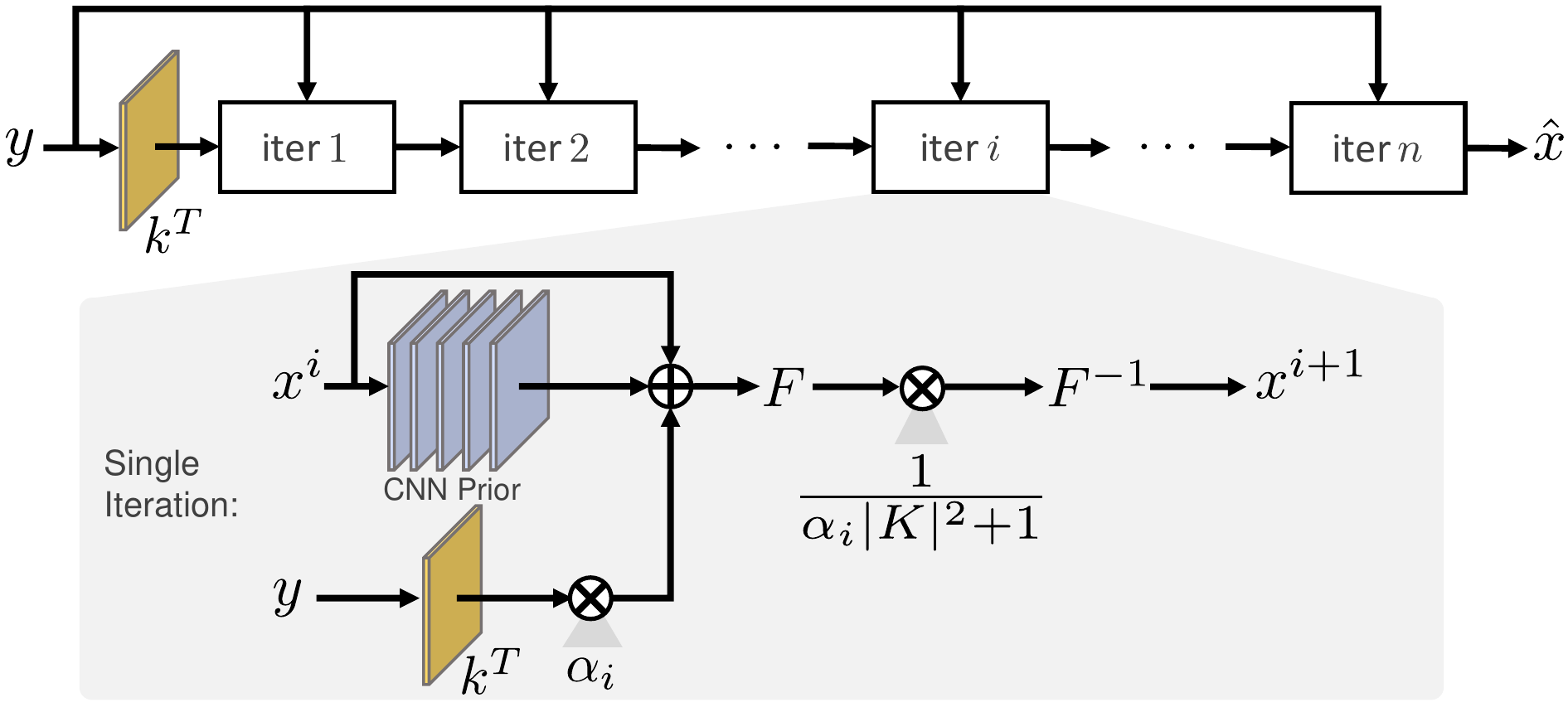}
\caption{A proximal gradient ODP network for deblurring under Gaussian noise,
  mapping the observation $y$ into an estimate $\hat{x}$ of the latent image $x$.
  Here $F$ is the DFT, $k$ is the blur kernel, and $K$ is its Fourier transform. }
\label{fig:optim_algs}
\vspace{-9pt}
\end{figure*}

Many efficient iterative optimization methods do not interact with the prior term $r$ directly,
but instead minimize $r$ via its (sub)gradient or proximal operator
$\prox{r(\cdot,\theta)}$, defined as
\[
\prox{r(\cdot,\theta)}(v) = \Argmin_z r(z,\theta) + \frac{1}{2}\|z - v\|^2_2.
\]
The proximal operator is a generalization of Euclidean projection.
In the ODP framework, we propose to parameterize the gradient or proximal operator of $r$
directly and define $r$ implicitly.

\section{Unrolled optimization with deep priors}\label{s-main}

We propose the ODP framework 
to incorporate knowledge of the image formation into deep convolutional networks.
The framework factors networks
into data steps, which are functions of the measurements $y$ encoding prior information about the image formation model,
and CNN steps, which represent statistical image priors.
The factorization follows a principled approach inspired by classical
optimization methods, thereby combining the best of deep models and classical algorithms.

\subsection{Framework}
The ODP framework is summarized by the network template in Algorithm~\ref{alg-meta}.
The design choices in the template are the optimization algorithm,
which defines the data step $\Gamma$ and algorithm state $z^k$,
the number of iterations $N$ that the algorithm is unrolled,
the function $\phi$ to initialize the algorithm from the measurements $y$, 
and the CNN used in the prior step,
whose output $x^{k+1/2}$ represents either $\nabla r(x^k,\theta^k)$ or
$\prox{r(\cdot,\theta^k)}(x^k)$,
depending on the optimization algorithm.
Figure~\ref{fig:optim_algs} shows an example ODP instance for deblurring under Gaussian noise.

\begin{algorithm}[h!]
\caption{ODP network template to solve problem (\ref{p-bayes})}\label{alg-meta}
\begin{algorithmic}[1]
\State{Initialization: $(x^0,z^0) = \phi(f,A,y,\theta^0)$.}
\For{$k=0$ to $N-1$}
\State $x^{k+1/2} \gets \textrm{CNN}(x^k,z^k,\theta^k)$.
\State $(x^{k+1},z^{k+1}) \gets \Gamma(f,A,y,x^{k+1/2},z^k,\theta^k)$.
\EndFor
\end{algorithmic}
\end{algorithm}

Instances of the ODP framework have two complementary interpretations.
From the perspective of classical optimization based methods,
an ODP architecture applies a standard optimization algorithm but learns a prior defined by a CNN.
From the perspective of deep learning, the network is a CNN with layers
tailored to the image formation model.

ODP networks are motivated by minimizing the objective in problem (\ref{p-bayes}), but they are trained to minimize a higher-level loss, which is defined on a metric between the network output and the ground-truth latent image over a training set of image/measurement pairs. Classical metrics for images are mean-squared error, PSNR, or SSIM. Let $\Gamma(y,\theta)$ be the network output given measurements $y$ and parameters $\theta$.
Then we train the network by (approximately) solving the optimization problem
\begin{equation}
\label{p-stoch}
\begin{array}{ll}
  \mbox{minimize} & E_{x \sim \Omega}E_{y \sim \omega(Ax)} \ell(x,\Gamma(y,\theta)),
\end{array}
\end{equation}
where $\theta$ is the optimization variable, $\ell$ is the chosen reconstruction loss, \eg, PSNR,
and $\Omega$ is the true distribution over images (as opposed to the
parameterized approximation $\Omega(\theta)$).
The ability to train directly for expected reconstruction loss is a primary advantage
of deep networks and unrolled optimization over classical image optimization methods. 
In contrast to pretraining priors on natural image training sets, 
directly training priors within the optimization algorithm allows ODP networks to learn application-specific priors and
efficiently share information between prior and data steps, allowing drastically
fewer iterations than classical methods.

Since ODPs are close to conventional CNNs,
we can approximately solve problem~(\ref{p-stoch}) using the many effective
stochastic gradient based methods developed for CNNs (\eg, Adam \cite{kingma2014adam}). 
Similarly, we can initialize the portions of $\theta$ corresponding to the CNN
prior steps using standard CNN initialization schemes (\eg, Xavier
initialization \cite{xavier}).
The remaining challenge in training ODPs is initializing the portions of $\theta$
corresponding to algorithm parameters in the data step $\Gamma$.
Most optimizaton algorithms only have one or two parameters per
data step, however,
so an effective initialization can be found through standard grid search.

\subsection{Design choices}\label{s-choices}
The ODP framework makes it straightforward to design a state-of-the-art network for solving an
inverse problem in imaging. The design choices are the choice of the optimization algorithm
to unroll, the CNN parameterization of the prior, and the initialization scheme.
In this section we discuss these design choices in detail and present
defaults guided by the empirical results in Section~\ref{s-results}.

\paragraph{Optimization algorithm}
The choice of optimization algorithm to unroll plays an important but poorly understood
role in the performance of unrolled optimization networks.
The only formal requirement is that each iteration of the unrolled algorithm
be almost everywhere differentiable.
Prior work has unrolled the proximal gradient method \cite{chen2015learning},
the half-quadratic splitting (HQS) algorithm \cite{schmidt2014shrinkage,geman1995nonlinear},
the alternating direction method of multipliers (ADMM) \cite{yang2016deep,Boyd:2011},
the Chambolle-Pock algorithm \cite{nips2016_6074,chambolle2011first},
ISTA \cite{gregor2010learning},
and a primal-dual algorithm with Bregman distances \cite{ochs2016techniques}.
No clear consensus has emerged as to which methods perform best in general
or even for specific problems.

In the context of solving problem~(\ref{p-bayes}),
we propose the proximal gradient method as a good default choice.
The method requires that the proximal operator of $g(x) = f(Ax,y)$
and its Jacobian can be computed efficiently.
Algorithm~\ref{alg-prox-grad} lists the ODP framework for the proximal
gradient method.
We interpret the CNN prior as $-\alpha_k\nabla r(x,\theta^k)$.
Note that for the proximal gradient network, the CNN prior is naturally a residual network
because its output $x^{k+1/2}$ is summed with its input $x^k$ in Step~\ref{pg-data-step}.

The algorithm parameters $\alpha_0,\ldots,\alpha_{N-1}$ represent the gradient
step sizes. The proposed initialization $\alpha_k = C_0 C^{-k}$ is based on an alternate
interpretation of Algorithm~\ref{alg-prox-grad} as an unrolled HQS method.
Adopting the aggressively decreasing $\alpha_k$ from
HQS minimizes the number of iterations needed \cite{geman1995nonlinear}.

\begin{algorithm}[h!]
\caption{ODP proximal gradient network.}\label{alg-prox-grad}
\begin{algorithmic}[1]
\State{Initialization: $x^0 = \phi(f,A,y,\theta^0)$, $\alpha_k = C_0 C^{-k}$, $C_0> 0$, $C > 0$.}
\For{$k=0$ to $N-1$}
\State $x^{k+1/2} \gets \textrm{CNN}(x^k,\theta^k)$.
\State $x^{k+1} \gets \Argmin_x \alpha_{k} f(Ax,y) +  \frac{1}{2}\|x -  x^k - x^{k+1/2}\|^2_2$.\label{pg-data-step}
\EndFor
\end{algorithmic}
\end{algorithm}

In Section~\ref{s-alg-compare}, we compare deblurring
and compressed sensing MRI results for ODP with proximal gradient, ADMM,
linearized ADMM (LADMM), and gradient descent.
The ODP formulations of ADMM, LADMM, and gradient descent can be found in the supplement.
We find that all algorithms that approximately invert the image formation
operator each iteration perform on par.
Algorithms such as ADMM and LADMM that incorporate Lagrange multipliers
were at best slightly better than simple primal algorithms like proximal
gradient and gradient descent for the low number of iterations typical for unrolled optimization methods.

\paragraph{CNN prior}
The choice of parameterizing each prior step as a separate CNN offers tremendous flexibility, even allowing the learning of a specialized function for each step. 
Algorithm \ref{alg-prox-grad} naturally introduces a residual connection to the
CNN prior, so a standard residual CNN is a reasonable default architecture choice.
The experiments in Section~\ref{s-results} show this architecture achieves
state-of-the-art results, while being easy to train with random initialization.

Choosing a CNN prior presents a trade-off between increasing the number of algorithm
iterations $N$, which adds alternation between data and prior steps, and making
the CNN deeper.
For example, in our experiments we found that for denoising, where the data step
is trivial, larger
CNN priors with fewer algorithm iterations gave better results,
while for deconvolution and MRI, where the data step is a complicated global operation, smaller priors and more iterations gave
better results.

\paragraph{Initialization}
The initialization function $(x^0,z^0) = \phi(f,A,y,\theta^0)$ could in theory be an arbitrarily complicated algorithm or neural network.
We found that the simple initialization $x^0 = A^Hy$, which is known as
backprojection, was sufficient for our applications \cite[Ch.~25]{Smith:1997}.

\section{Related work}\label{s-prev-work}

The ODP framework generalizes and improves upon previous work on unrolled
optimization and deep models for inverse imaging problems.

\paragraph{Unrolled optimization networks}
An immediate choice in constructing unrolled optimization networks for inverse
problems in imaging is to parameterize the prior $r(x,\theta)$ as $r(x,\theta) =
\|Cx\|_1$,
where $C$ is a filterbank, meaning a linear operator given by $Cx = (c_1 * x, \ldots, c_p * x)$
for convolution kernels $c_1,\ldots,c_p$.
The parameterizaton is inspired by
hand-engineered priors that exploit the sparsity of images in a particular
(dual) basis, such as anisotropic total-variation.
Unrolled optimization networks with learned sparsity priors have achieved good
results but are limited in representational power by the choice of a simple $\ell_1$-norm \cite{gregor2010learning,nips2016_6074}.

\paragraph{Field-of-experts}
A more sophisticated approach than learned sparsity priors is to parameterize the prior
gradient or proximal operator as a field-of-experts (FoE) $g(Cx,\theta)$,
where $C$ is again a filterbank and $g$ is a separable nonlinearity parameterized by $\theta$,
such as a sum of radial basis functions \cite{roth2005fields, schmidt2014shrinkage, chen2015learning, yang2016deep}.
The ODP framework improves upon the FoE approaches both empirically,
as shown in the model comparisons in Section \ref{s-results},
and theoretically, as the FoE model is essentially a 2-layer CNN and so 
has less representational power than deeper CNN priors.


\paragraph{Deep models for direct inversion}
Several recent approaches propose CNNs that directly solve specific imaging problems.
These architectures resemble instances of the ODP framework,
though with far different motivations for the design.
Schuler et al.~propose a network for deblurring that applies a single, fixed deconvolution step followed by a learned CNN,
akin to a prior step in ODP with a different initial iterate \cite{Schuler_2013_CVPR}. 
Xu et al.~propose a network for deblurring that applies a single, learned deconvolution
step followed by a CNN, similar to a one iteration ODP network \cite{xu2014deep}.
Wang et al.~propose a CNN for MRI whose output is
averaged with observations in $k$-space,
similar to a one iteration ODP network but without jointly learning the prior
and data steps \cite{wang:16}.
We improve on these deep models by recognizing the connection to classical
optimization based methods via the ODP framework and using the framework
to design more powerful, multi-iteration architectures.

\section{Experiments}\label{s-results}

\begin{figure*}
\begin{center}
\includegraphics[width=\textwidth]{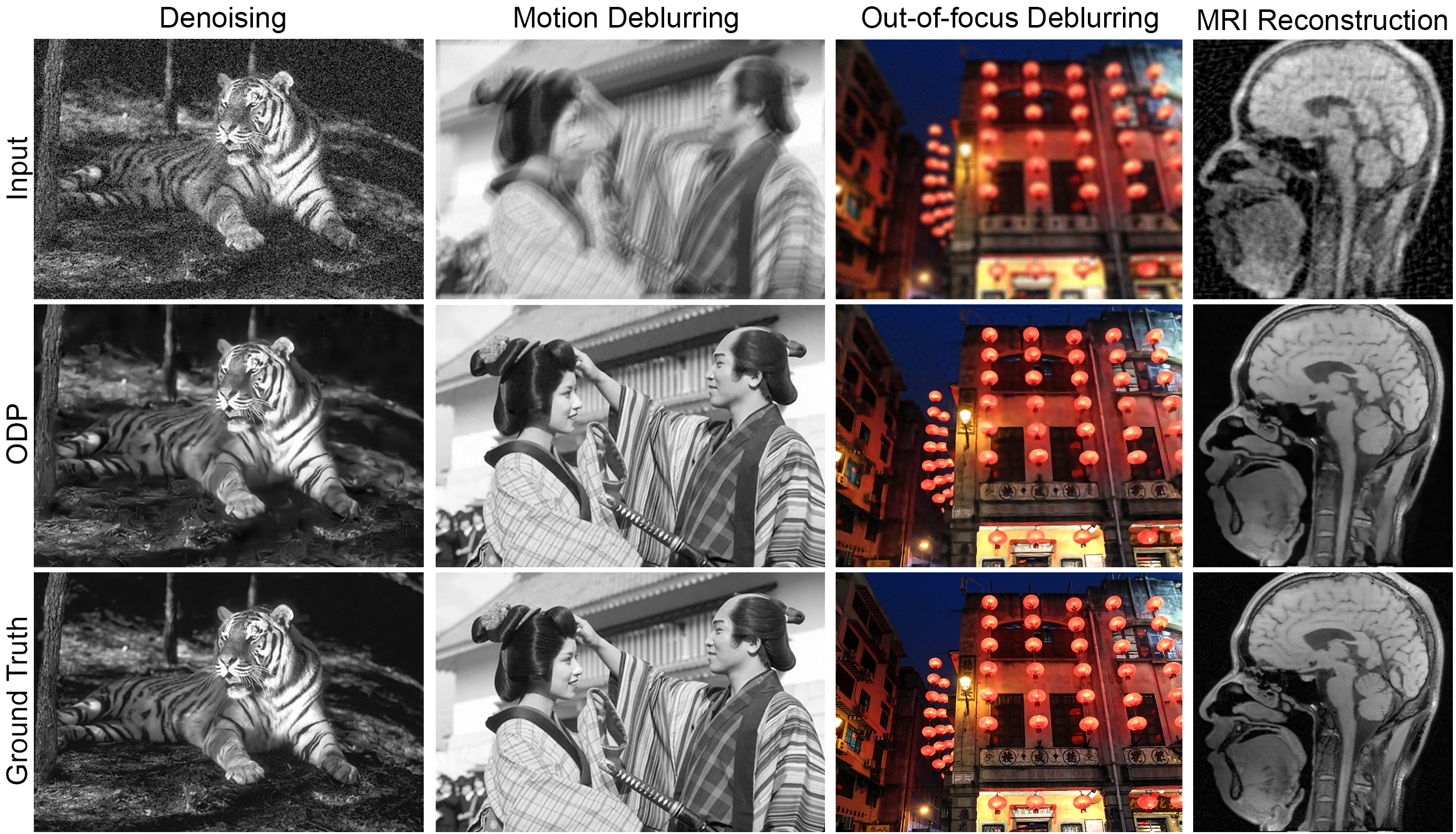}
\end{center}
\caption{A qualitative overview of results with ODP networks.}\label{fig-qual}
\vspace{-4pt}
\end{figure*}

In this section we present results and analysis of ODP networks for denoising, deblurring, and compressed sensing MRI.
Figure \ref{fig-qual} shows a qualitative overview for this variety of inverse imaging problems.
Please see the supplement for details of the training procedure for each experiment.

\subsection{Denoising}\label{s-denoise}
We consider the Gaussian denoising problem with image formation $y = x + z$,
with $z \sim \mathcal{N}(0,\sigma^2)$. The corresponding Bayesian estimation
problem (\ref{p-bayes}) is 
\begin{equation}
\label{eq:deconv}
\begin{array}{ll}
\mbox{minimize} & \frac{1}{2\sigma^2}\|x - y\|^2_2 + r(x,\theta).
\end{array}
\end{equation}
We trained a 4 iteration proximal gradient ODP network with a 10 layer, 64 channel
residual CNN prior on the 400 image training set from \cite{nips2016_6074}.
Table \ref{tab-denoise} shows that the ODP network outperforms all state-of-the-art
methods on the 68 image test set evaluated in \cite{nips2016_6074}.
%
%
\begin{table}
\parbox{.475\linewidth}{
\centering
\caption{Average PSNR (dB) for noise level and test set from
  \cite{nips2016_6074}.
Timings were done on Xeon 3.2 GHz CPUs and a Titan X GPU.}
\vspace{\baselineskip}
\label{tab-denoise}
\begin{tabular}{lcl}
\hline
  Method & $\sigma = 25$ & Time per image \\ \hline
  BM3D \cite{dabov2007image} & 28.56 & 2.57s (CPU) \\
  EPLL \cite{zoran2011learning} & 28.68 & 108.72s (CPU) \\
  Schmidt \cite{schmidt2014shrinkage} & 28.72 & 5.10s (CPU) \\
  Wang \cite{nips2016_6074} & 28.79 & 0.011s (GPU) \\
  Chen \cite{chen2015learning} & 28.95 & 1.42s (CPU) \\
  \textbf{ODP} & \textbf{29.04} & 0.13s (GPU) 
\end{tabular}
}
\hfill
\parbox{.475\linewidth}{
\centering
\vspace{-12pt}
\caption{Average PSNR (dB) for kernels and test set from \cite{xu2014deep}.}
\label{tab-xu}
\vspace{\baselineskip}
\begin{tabular}{ccccccccc}
\hline
  Method & Disk & Motion \\ \hline
  Krishnan \cite{krishnan2009fast} & 25.94 & 25.07 \\
  Levin \cite{Levin:07} & 24.54 & 24.47 \\
  Schuler \cite{Schuler_2013_CVPR} & 24.67 & 25.27 \\
  Schmidt \cite{schmidt2014shrinkage} & 24.71 & 25.49 \\
  Xu \cite{xu2014deep}  & 26.01 & 27.92 \\
  \textbf{ODP} & \textbf{26.11} & \textbf{28.49}
\end{tabular}
}
\vspace{-10pt}
\end{table}

\subsection{Deblurring}
We consider the problem of joint Gaussian denoising and deblurring,
in which the latent image $x$ is convolved with a known blur kernel
in addition to being corrupted by Gaussian noise.
The image formation model is $y = k*x + z$, where
$k$ is the blur kernel and $z \sim \mathcal{N}(0,\sigma^2)$.
The corresponding Bayesian estimation problem~(\ref{p-bayes}) is
\vspace{-5pt}
\begin{equation}
\label{p-deconv}
\begin{array}{ll}
\mbox{minimize} & \frac{1}{2\sigma^2}\|k*x - y\|^2_2 + r(x,\theta).
\end{array}
\end{equation}
To demonstrate the benefit of ODP networks specialized to specific problem instances,
we first train models per kernel, as proposed in \cite{xu2014deep}. Specifically, we train 8 iteration proximal gradient ODP networks with residual 5 layer, 64
channel CNN priors for the out-of-focus disk kernel and motion blur kernel from \cite{xu2014deep}.
Following Xu et al., we train one model per kernel on ImageNet \cite{deng2009imagenet}, including clipping and JPEG artifacts as required by the authors. 
Table \ref{tab-xu} shows that the ODP networks outperform prior work slightly on
the low-pass disk kernel, which completely removes high frequency content, while
a substantial gain is achieved for the motion blur kernel, which preserves
more frequency content and thus benefits from the inverse image formation steps in ODP.

Next, we show that ODP networks can generalize across image formation models.
Table \ref{tab-schuler} compares ODP networks (same architecture as above) on
the test scenarios from~\cite{Schuler_2013_CVPR}. We trained one ODP network for
the four out-of-focus kernels and associated noise levels in
\cite{Schuler_2013_CVPR}. The out-of-focus model is on par with of Schuler et
al., even though Schuler et al.~train a specialized model for each kernel and
associated noise level. We trained a second ODP network on randomly generated
motion blur kernels. This model outperforms Schuler et al.~on the unseen test
set motion blur kernel, even though Schuler et al.~trained specifically on the motion blur kernel from their test set.
%
%
\begin{table}
\centering
\caption{Average PSNR (dB) for kernels and test set of 11 images from
  \cite{Schuler_2013_CVPR}.
  We also evaluate Schuler et al.~and ODP on the BSDS500 data set for a lower variance
  comparison \cite{MartinFTM01}.}
\label{tab-schuler}
\vspace{\baselineskip}
\begin{tabular}{lccccc}
\hline
  Model & Gaussian (a) & Gaussian (b) & Gaussian (c) & Box (d) & Motion (e) \\
        & $\sigma=10.2$ & $\sigma=2$ & $\sigma=10.2$ & $\sigma=2.5$ &
                                                                      $\sigma=2.5$
  \\
\hline
  EPLL \cite{zoran2011learning} & 24.04 & 26.64 & 21.36 & 21.04 & 29.25 \\
  Levin \cite{Levin:07} & 24.09 & 26.51 & 21.72 & 21.91 & 28.33 \\
 Krishnan \cite{krishnan2009fast} & 24.17 & 26.60 & 21.73 & 22.07 & 28.17 \\
  IDD-BM3D \cite{danielyan2012bm3d} & 24.68 & 27.13 & 21.99 & 22.69 & 29.41 \\
  Schuler \cite{Schuler_2013_CVPR} & \textbf{24.76/26.19} & \textbf{27.23/28.41}
& \textbf{22.20/24.06} & \textbf{22.75/24.37} & 29.42/29.91 \\
\textbf{ODP} & \textbf{24.89/26.20} & \textbf{27.44/28.44} &
\textbf{22.26/23.97} &   \textbf{23.17/24.37} & \textbf{30.54/30.50}
                                                                         \end{tabular}
\vspace{-8pt}
\end{table}

\subsection{Compressed sensing MRI}
In compressed sensing (CS) MRI, a latent image $x$ is measured in the Fourier
domain with subsampling. Following~\cite{yang2016deep}, we assume noise free measurements.
The image formation model is $y = P\mathcal{F}x$, where
$\mathcal{F}$ is the DFT and $P$ is a diagonal binary sampling matrix for a given subsampling pattern.
The corresponding Bayesian estimation problem~(\ref{p-bayes}) is
\begin{equation}
\label{p-mri}
\begin{array}{ll}
\mbox{minimize} & r(x,\theta) \\
\mbox{subject to} & P\mathcal{F}x = y.
\end{array}
\end{equation}
We trained an 8 iteration proximal gradient ODP network with a residual 7 layer, 64 channel CNN prior on the 100 training images and pseudo-radial sampling pattern from~\cite{yang2016deep}, range from sampling 20\% to 50\% of the Fourier domain. We evaluate on the 50 test images from the same work.
%
\begin{table}[t]
\caption{Average PSNR (dB) on the brain MRI test set from \cite{yang2016deep}
  for different sampling ratios. Timings were done on an Intel i7-4790k 4 GHz
  CPU and Titan X GPU.}
\label{tab-mri}
\vspace{\baselineskip}
\centering
\begin{tabular}{lccccl}
\hline
Method & 20\% & 30\% & 40\% & 50\% & Time per image \\ \hline
PBDW \cite{qu2012undersampled} & 36.34 & 38.64 & 40.31 & 41.81 & 35.36s (CPU) \\
PANO \cite{qu2014magnetic} & 36.52 & 39.13 & 41.01 & 42.76 & 53.48s (CPU) \\
FDLCP \cite{zhan2016fast} & 36.95 & 39.13 & 40.62 & 42.00 & 52.22s (CPU) \\
ADMM-Net \cite{yang2016deep} & 37.17 & 39.84 & 41.56 & 43.00 & 0.79s (CPU) \\
BM3D-MRI \cite{Eksioglu2016} & 37.98 & 40.33 & 41.99 & 43.47 & 40.91s (CPU)\\
\textbf{ODP} & \textbf{38.50} & \textbf{40.71} & \textbf{42.34} & \textbf{43.85}
                                   & 0.090s (GPU)
\end{tabular}
\vspace{-10pt}
\end{table}
Table \ref{tab-mri} shows that the ODP network outperforms even the the best alternative method,
BM3D-MRI, for all sampling patterns. The improvement over BM3D-MRI is larger for sparser (and thus more challenging) sampling patterns.
The fact that a single ODP model performed so well on all four sampling
patterns, particularly in comparison to the ADMM-Net models, which were trained separately per sampling pattern,
again demonstrates that ODP generalizes across image formation models.

\subsection{Contribution of prior information}\label{s-resid}
In this section, we analyze how much of the reconstruction performance is due to
incorporating the image formation model in the data step and how much is due to
the CNN prior steps.
To this end, we performed an ablation study where ODP models were trained
without the data steps.
The resulting models are pure residual networks.
\begin{table*}[!htb]
\vspace{-2pt}
\centering
\begin{minipage}{1.0\linewidth}
Average PSNR (dB) for the noise level $\sigma=25$ on the test set from \cite{nips2016_6074},
  over 8 unseen motion blur kernels from \cite{Levin:07} for the noise level
  $\sigma=2.55$ on the BSDS500 \cite{MartinFTM01}, 
  and over the MRI sampling patterns and test set from \cite{yang2016deep}.
	\vspace{\baselineskip}
\end{minipage} 
    \begin{minipage}{.5\linewidth}
\caption{Performance with and without data step.}
\label{tab-prior-info}
\vspace{4pt}
      \centering
\begin{tabular}{lcc}
\hline
Application & Proximal gradient & Prior only\\ \hline
Denoising & 29.04 & 29.04 \\
Deblurring & 31.04 & 26.04 \\
CS MRI & 41.35 & 37.70 
\end{tabular}
    \end{minipage}%
    \begin{minipage}{.5\linewidth}
				\vspace{-10pt}
\caption{Comparison of different algorithms.}
\label{tab-algs}
\vspace{4pt}
      \centering
\begin{tabular}{lccc}
\hline
Model & Deblurring & CS MRI \\ \hline
Proximal gradient & 31.04 &  41.35 \\
ADMM & 31.01 & 41.39 \\
LADMM & 30.17 & 41.37  \\
Gradient descent & 29.96 & N/A
\end{tabular}
\vspace{-3pt}
    \end{minipage} 
		\vspace{-13pt}
\end{table*}



Table \ref{tab-prior-info} shows that for denoising the residual network performed as well as the proximal
gradient ODP network,
while for deblurring and CS MRI the proximal gradient network performed
substantially better.
The difference between denoising and the other two inverse problems is that in
the case of denoising the ODP data step is trivial: inverting $A=I$.
However, for deblurring the data step applies $B = (A^HA + \gamma I)^{-1}A^H$, a complicated global operation that for motion blur satisfies $BA \approx I$. Similarly, for CS MRI the ODP proximal gradient network applies $A^\dagger$ in the data step, a complicated global operation (involving a DFT) that satisfies $BA \approx I$.

The results suggest an interpretation of the ODP proximal gradient networks
as alternating between applying local corrections in the CNN
prior step and applying a global operator $B$ such that $BA \approx I$ in the data step.
The CNNs in the ODP networks conceptually learn to denoise and correct errors introduced
by the approximate inversion of $A$ with $B$,
whereas the pure residual networks must learn to denoise and invert $A$ directly.
When approximately inverting $A$ is a complicated global operation, direct inversion using residual networks poses an extreme challenge, overcome by the indirect approach taken by the ODP networks.

\subsection{Comparing algorithms}\label{s-alg-compare}
The existing work on unrolled optimization has done little to clarify which
optimization algorithms perform best when unrolled.
We investigate the relative performance of unrolled optimization algorithms in the context of
ODP networks. We have compared ODP networks for ADMM, LADMM, proximal gradient
and gradient descent in Table~\ref{tab-algs}, using the same initialization and
network parameters as for deblurring and CS MRI.

For deblurring, the proximal gradient and ADMM models, which apply the regularized
pseudoinverse $(A^HA + \gamma I)^{-1}A^H$ in the data step,
outperform the LADMM and gradient descent models, which only apply $A$ and $A^H$.
The results suggest that taking more aggressive steps to approximately invert
$A$ in the data step improves performance.
For CS MRI, all algorithms apply the pseudoinverse $A^\dagger$ (because
$A^H=A^\dagger$) and have similar performance,
which matches the observations for deblurring.

Our algorithm comparison shows minimal benefits to incorporating Lagrange multipliers, which is expected for the relatively low number of iterations $N = 8$ in our models.
The ODP ADMM networks for deblurring and CS MRI are identical to the respective
proximal gradient networks except that ADMM includes Lagrange multipliers,
and the performance is on par.
For deblurring, LADMM and gradient descent are similar architectures,
but LADMM incorporates Lagrange multipliers and shows a small performance gain.
Note that Gradient descent cannot be applied to CS MRI as problem~(\ref{p-mri}) is constrained.


\section{Conclusion}
The proposed ODP framework offers a principled approach to incorporating prior knowledge of the image formation into deep networks for solving inverse problems in imaging,
yielding state-of-the-art results for denoising, deblurring, and CS MRI.
The framework generalizes and outperforms previous approaches to unrolled optimization and deep networks for direct inversion. The presented ablation studies offer general insights into
the benefits of prior information and what algorithms are most suitable for unrolled optimization.

Although the class of imaging problems considered in this work lies at the core of imaging and sensing, it is only a
small fraction of the potential applications for ODP.
In future work we will explore ODP for blind inverse problems,
in which the image formation operator $A$ is not fully known,
as well as nonlinear image formation models.
Outside of imaging, control is a promising field in which to apply the
ODP framework because deep networks may potentially benefit from prior knowledge of
physical dynamics.




\small

\bibliographystyle{plain}
\bibliography{main}

\begin{thebibliography}{10}

\bibitem{doi:10.1137/080716542}
A.~Beck and M.~Teboulle.
\newblock A fast iterative shrinkage-thresholding algorithm for linear inverse
  problems.
\newblock {\em SIAM Journal on Imaging Sciences}, 2(1):183--202, 2009.

\bibitem{Boyd:2011}
S.~Boyd, N.~Parikh, E.~Chu, B.~Peleato, and J.~Eckstein.
\newblock Distributed optimization and statistical learning via the alternating
  direction method of multipliers.
\newblock {\em Foundations and Trends in Machine Learning}, 3(1):1--122, 2001.

\bibitem{boyd2004cvx}
S.~Boyd and L.~Vandenberghe.
\newblock {\em Convex Optimization}.
\newblock Cambridge University Press, 2004.

\bibitem{chambolle2011first}
A.~Chambolle and T.~Pock.
\newblock A first-order primal-dual algorithm for convex problems with
  applications to imaging.
\newblock {\em Journal of Mathematical Imaging and Vision}, 40(1):120--145,
  2011.

\bibitem{chen2015learning}
Y.~Chen, W.~Yu, and T.~Pock.
\newblock On learning optimized reaction diffusion processes for effective
  image restoration.
\newblock In {\em Proceedings of the IEEE Conference on Computer Vision and
  Pattern Recognition}, pages 5261--5269, 2015.

\bibitem{dabov2007image}
K.~Dabov, A.~Foi, V.~Katkovnik, and K.~Egiazarian.
\newblock Image denoising by sparse {3-D} transform-domain collaborative
  filtering.
\newblock {\em IEEE Trans. Image Processing}, 16(8):2080--2095, 2007.

\bibitem{danielyan2012bm3d}
A.~Danielyan, V.~Katkovnik, and K.~Egiazarian.
\newblock {BM3D} frames and variational image deblurring.
\newblock {\em IEEE Trans. Image Processing}, 21(4):1715--1728, 2012.

\bibitem{deng2009imagenet}
Jia Deng, Wei Dong, Richard Socher, Li-Jia Li, Kai Li, and Li~Fei-Fei.
\newblock Imagenet: A large-scale hierarchical image database.
\newblock In {\em Computer Vision and Pattern Recognition, 2009. CVPR 2009.
  IEEE Conference on}, pages 248--255. IEEE, 2009.

\bibitem{Eksioglu2016}
E.~Eksioglu.
\newblock Decoupled algorithm for {MRI} reconstruction using nonlocal block
  matching model: {BM3D-MRI}.
\newblock {\em Journal of Mathematical Imaging and Vision}, 56(3):430--440,
  2016.

\bibitem{geman1995nonlinear}
D.~Geman and C.~Yang.
\newblock Nonlinear image recovery with half-quadratic regularization.
\newblock {\em IEEE Transactions on Image Processing}, 4(7):932--946, 1995.

\bibitem{xavier}
X.~Glorot and Y.~Bengio.
\newblock Understanding the difficulty of training deep feedforward neural
  networks.
\newblock In {\em Proceedings of the International Conference on Artificial
  Intelligence and Statistics}, volume~9, pages 249--256, 2010.

\bibitem{gregor2010learning}
K.~Gregor and Y.~LeCun.
\newblock Learning fast approximations of sparse coding.
\newblock In {\em Proceedings of the International Conference on Machine
  Learning}, pages 399--406, 2010.

\bibitem{kingma2014adam}
D.~Kingma and J.~Ba.
\newblock Adam: A method for stochastic optimization.
\newblock {\em arXiv preprint arXiv:1412.6980}, 2014.

\bibitem{krishnan2009fast}
D.~Krishnan and R.~Fergus.
\newblock Fast image deconvolution using hyper-{L}aplacian priors.
\newblock In {\em Advances in Neural Information Processing Systems}, pages
  1033--1041, 2009.

\bibitem{Levin:07}
A.~Levin, R.~Fergus, F.~Durand, and W.~Freeman.
\newblock Image and depth from a conventional camera with a coded aperture.
\newblock {\em ACM Transactions on Graphics}, 26(3), 2007.

\bibitem{MartinFTM01}
D.~Martin, C.~Fowlkes, D.~Tal, and J.~Malik.
\newblock A database of human segmented natural images and its application to
  evaluating segmentation algorithms and measuring ecological statistics.
\newblock In {\em Proceedings of the IEEE International Conference on Computer
  Vision}, pages 416--423, 2001.

\bibitem{ochs2016techniques}
p.~Ochs, R.~Ranftl, T.~Brox, and T.~Pock.
\newblock Techniques for gradient-based bilevel optimization with non-smooth
  lower level problems.
\newblock {\em Journal of Mathematical Imaging and Vision}, pages 1--20, 2016.

\bibitem{qu2012undersampled}
X.~Qu, D.~Guo, B.~Ning, Y.~Hou, Y.~Lin, S.~Cai, and Z.~Chen.
\newblock Undersampled {MRI} reconstruction with patch-based directional
  wavelets.
\newblock {\em Magnetic Resonance Imaging}, 30(7):964--977, 2012.

\bibitem{qu2014magnetic}
X.~Qu, Y.~Hou, F.~Lam, D.~Guo, J.~Zhong, and Z.~Chen.
\newblock Magnetic resonance image reconstruction from undersampled
  measurements using a patch-based nonlocal operator.
\newblock {\em Medical Image Analysis}, 18(6):843--856, 2014.

\bibitem{roth2005fields}
S.~Roth and M.~Black.
\newblock Fields of experts: A framework for learning image priors.
\newblock In {\em Proceedings of the IEEE Conference on Computer Vision and
  Pattern Recognition}, pages 860--867, 2005.

\bibitem{schmidt2014shrinkage}
U.~Schmidt and S.~Roth.
\newblock Shrinkage fields for effective image restoration.
\newblock In {\em Proceedings of the IEEE Conference on Computer Vision and
  Pattern Recognition}, pages 2774--2781, 2014.

\bibitem{Schuler_2013_CVPR}
C.~Schuler, H.~Burger, S.~Harmeling, and B.~Scholkopf.
\newblock A machine learning approach for non-blind image deconvolution.
\newblock In {\em Proceedings of the IEEE Conference on Computer Vision and
  Pattern Recognition}, pages 1067--1074, 2013.

\bibitem{Smith:1997}
S.~Smith.
\newblock {\em The Scientist and Engineer's Guide to Digital Signal
  Processing}.
\newblock California Technical Publishing, San Diego, 1997.

\bibitem{nips2016_6074}
S.~Wang, S.~Fidler, and R.~Urtasun.
\newblock Proximal deep structured models.
\newblock In {\em Advances in Neural Information Processing Systems 29}, pages
  865--873. 2016.

\bibitem{wang:16}
S.~Wang, Z.~Su, L.~Ying, X.~Peng, S.~Zhu, F.~Liang, D.~Feng, and D.~Liang.
\newblock Accelerating magnetic resonance imaging via deep learning.
\newblock In {\em Proceedings of the IEEE International Symposium on Biomedical
  Imaging}, pages 514--517, 2016.

\bibitem{xu2014deep}
L.~Xu, J.~Ren, C.~Liu, and J.~Jia.
\newblock Deep convolutional neural network for image deconvolution.
\newblock In {\em Advances in Neural Information Processing Systems}, pages
  1790--1798, 2014.

\bibitem{yang2016deep}
Y.~Yang, J.~Sun, H.~Li, and Z.~Xu.
\newblock Deep {ADMM-N}et for compressive sensing {MRI}.
\newblock In {\em Advances in Neural Information Processing Systems}, pages
  10--18, 2016.

\bibitem{zhan2016fast}
Z.~Zhan, J.-F. Cai, D.~Guo, Y.~Liu, Z.~Chen, and X.~Qu.
\newblock Fast multiclass dictionaries learning with geometrical directions in
  {MRI} reconstruction.
\newblock {\em IEEE Transactions on Biomedical Engineering}, 63(9):1850--1861,
  2016.

\bibitem{zoran2011learning}
D.~Zoran and Y.~Weiss.
\newblock From learning models of natural image patches to whole image
  restoration.
\newblock In {\em Proceedings of the IEEE International Conference on Computer
  Vision}, pages 479--486, 2011.

\end{thebibliography}

\end{document}